\documentclass[fleqn,10pt]{wlscirep}
\usepackage[utf8]{inputenc}
\usepackage[T1]{fontenc}
\usepackage{multirow}
\usepackage{subcaption,graphicx}
\usepackage{float}
\usepackage{xcolor,cancel}
\usepackage{color}
\usepackage{xcolor}
\usepackage[normalem]{ulem}
\usepackage{soul}

\title{Feasibility Study of Multi-Site Split Learning for Privacy-Preserving Medical Systems under Data Imbalance Constraints in COVID-19, X-Ray, and Cholesterol Dataset}

\author[1]{Yoo Jeong Ha}
\author[1]{Gusang Lee}
\author[1]{Minjae Yoo}
\author[2,*]{Soyi Jung}
\author[3,**]{Seehwan Yoo}
\author[1,***]{Joongheon Kim}
\affil[1]{Korea University, School of Electrical Engineering, Seoul, 02841, Republic of Korea}
\affil[2]{Hallym University, School of Software, Chuncheon, 24252, Republic of Korea}
\affil[3]{Dankook University, Department of Mobile Systems Engineering, Yongin, 16890, Republic of Korea}

\affil[*]{sjung@hallym.ac.kr}
\affil[**]{seehwan.yoo@dankook.ac.kr}
\affil[***]{joongheon@korea.ac.kr}

\begin{abstract}
It seems as though progressively more people are in the race to upload content, data, and information online; and hospitals haven’t neglected this trend either. Hospitals are now at the forefront for multi-site medical data sharing to provide groundbreaking advancements in the way health records are shared and patients are diagnosed. Sharing of medical data is essential in modern medical research. Yet, as with all data sharing technology, the challenge is to balance improved treatment with protecting patient’s personal information. This paper provides a novel split learning algorithm coined the term, “multi-site split learning”, which enables a secure transfer of medical data between multiple hospitals without fear of exposing personal data contained in patient records. It also explores the effects of varying the number of end-systems and the ratio of data-imbalance on the deep learning performance. A guideline for the most optimal configuration of split learning that ensures privacy of patient data whilst achieving performance is empirically given. We argue the benefits of our multi-site split learning algorithm, especially regarding the privacy preserving factor, using CT scans of COVID-19 patients, X-ray bone scans, and cholesterol level medical data.  
\end{abstract}
\begin{document}

\flushbottom
\maketitle
\thispagestyle{empty}

% \noindent Please note: Abbreviations should be introduced at the first mention in the main text – no abbreviations lists. Suggested structure of main text (not enforced) is provided below.

\section*{Introduction}
% It is no secret that the world is becoming an ever more integrated society through the advent of the Internet. 

%  thrum of concern over the so-called Delta variant grows steadily louder.
% forcing many countries to reimpose stringent restrictions on social activity.
% where wide swaths of the population are immunized, the Delta variant has outpaced vaccination efforts, pushing the goal of herd immunity further out of reach and postponing an end to the pandemic.experience is a harbinger of what’s to come, less effective in curtailing Delta’s spread 
% helped vaccinate more than 800 million children against pneumonia,

The world was met with a demise when the lethal Coronavirus Disease 2019 (COVID-19) began taking away the lives of loved ones. COVID-19 has infected 221,648,869 and killed 4,582,338 people worldwide as of September 8th, 2021\cite{WHO}. Since both case numbers and deaths linked to COVID-19 continue to rise as new and more aggressive variants of the contagion reach the population, the world is shifting further away from the end of the COVID-19 pandemic. This ongoing global pandemic has divided the world apart at a time where mankind should unite to fight this fatal disease together. At the heart of this step to amalgamate the deeply detached, is our proposed “multi-site split learning” algorithm. This paper aims to furnish a secure learning process where hospitals all over the globe can share their findings to create a deep learning model without revealing any sensitive patient information. This new technique allows hospitals to share their data without sending raw medical images to external associations. Our privacy-enhancing technology provides experimental results using real-world data, with only an encrypted image being divulged to an external server.

The global health crisis has led to countless research introducing new technology that can aid in the battle against the deadly disease. Some papers such as\cite{ABDELBASSET2021106647}  have introduced an efficient segmentation of 2019-nCov infection (FSS-2019-nCov) from a limited source of annotated CT scans of the lungs. Since the outburst, endless research has been conducted in comparing the performance of different CNN models \cite{dhiman2021adopt, ismael2021deep, sethy2021computer, rahman2021exploring, emara2021deep}  for the detection of patients infected with COVID-19. Paper \cite{9343340}  has even integrated extended reality (XR) with deep-learning-based Internet-of-Medical-Things (IoMT) for a telemedicine diagnosis of COVID-19. While most studies have actively sought to test which model is the best fit for the classification of the patients with and without the virus, these papers do not touch on the privacy aspect of patient data. The aforementioned papers do not explore methods that protect the sensitive information that entails the CT scans or X-ray images of patients.
 Other split learning algorithms have been introduced before as well \cite{vepakomma2018split, thapa2020splitfed}.  These papers compared results with federated learning, another form of distributed machine learning, and affirmed the benefits of collaborative training of deep neural networks in health care. Yet, these papers do not extend to the multiple client element of split learning. Allowing multiple hospitals to share their findings and collaborate to build a global COVID-19 classification model turns over a new leaf for the split learning algorithm. Furthermore, \cite{ha2021spatio}  explores 3-client split learning but does not examine the variation to the number of participating clients. No other paper explores the effect of changing the data split ratio on the performance of split learning. 
Hence, the main contributions of our paper are:
\begin{itemize}
    \item  This paper explores the optimal number of clients for multi-site split learning through experimental analysis.  
    \item  This research focuses on empirically delivering the optimal data split ratio for the best split learning performance. 
    \item  This study is conducted using a real-world cholesterol dataset from Seoul National University Hospital (SNUH), and other open datasets--COVID-19 chest CT scans and MURA bone X-ray scans.  
\end{itemize}

 The main research goal of this paper is to test the effect of varying the number of participating hospitals, which are called end-systems, and the data ratio on the performance of our multi-site split learning algorithm. As more citizens are hospitalized due to the COVID-19 virus, a precise identification tool in the hospital to help accelerate the treatment process is imperative. Through this paper's experimental findings on the optimal client number and data split ratio, we hope researchers would use our guidance to hastily practice split learning while protecting the personal information of patient records. 

\typeout{In split learning, the learning process is, as the word suggests, literally split or separated into two parts: the end-systems and the server. Here, the end-systems refer to the multiple hospitals that possess original medical data to be trained in the deep learning model. The server is where the actual learning of the deep neural network occurs. Multiple hospitals are connected to one server; all the hospitals connected to the server collaborate to train the very deep neural network that is placed in the server without exposing their raw data to an external network. A meticulously devised framework of our proposed algorithm is shown in Figure \ref{fig:framework}. Every individual hospital only runs the deep neural network up to the first hidden layer. After passing through the first hidden layer, the end-systems send only their feature maps to the server. Since only the encrypted feature map is transferred to the external network, and not the original medical data, the privacy of confidential patient information is preserved. It is coined "multi-site split learning" since multiple hospitals (multi-site) are geometrically spaced apart from each other and the deep neural network is divided among them.}

\typeout{As more citizens are hospitalized due to the COVID-19 virus, a precise identification tool in the hospital to help accelerate the treatment process is imperative. }This powerful tool can be utilized with various types of data such as image data or numerical data. With the appropriate kind of deep learning model selected, our algorithm is applicable to accompany a wide variety of data types. It has been reported that COVID-19 causes severe acute respiratory syndrome coronavirus 2 (SARA-CoV-2). With the deployment of our novel technique, a precise classification of pneumonia caused by the COVID-19 virus can be made. This dataset is utilized in such a timely manner for this pandemic; this technological assistance can serve as a means to slowing down the number of patients getting infected with SARA-CoV-2 by devising initial treatment plans. Also, we use MUsculoskeletal RAdiographs (MURA) datasets, a collection of X-ray scans from seven body parts, to classify whether a bone is fractured or not. According to \cite{briggs2018reducing}, the musculoskeletal condition affects one in two American adults, which accounts for far more than that of cardiovascular and chronic respiratory diseases combined. Hence, this dataset is used to validate our model and aid in treating this common illness. Since our algorithm can be applied to different data types, we also use numerical data--cholesterol levels--to predict the level of harmful cholesterol using other patient information. We use three medical datasets to corroborate our claim on the benefits of our multi-site split learning algorithm. 

% swiftly devise treatment plans  

% hosiptals more resources can be redirected to searc
% The desire for privacy is a universal trait across time and cultures. 

% The Introduction section, of referenced text\cite{Figueredo:2009dg} expands on the background of the work (some overlap with the Abstract is acceptable). The introduction should not include subheadings.

% \begin{figure}[ht]
% \centering
% \includegraphics[width=1.10\linewidth]{images/fig1_real.pdf}
% \caption{The overall framework of our multi-site split learning algorithm.}
% \label{fig:framework}
% \end{figure}

% \begin{figure}[ht]
% \centering
% \includegraphics[width=0.66\linewidth]{images/nature_fig2.pdf}
% \caption{A generalized deep learning network for COVID-19 classification.}
% \label{fig:deeplearning}
% \end{figure}

\section*{Methods}

% Topical subheadings are allowed. Authors must ensure that their Methods section includes adequate experimental and characterization data necessary for others in the field to reproduce their work.

\subsection*{\typeout{Data Collection or }Study Datasets}
We use two different types of data--image and numerical data to give credence to our multi-site split learning algorithm. COVID-19 chest computed tomography (CT) scans and MURA bone X-ray images are used for classification and cholesterol data, a numerical data, is analyzed for prediction. To enhance diagnoses and speed up treatment options for patients in the time of this global epidemic, our original split learning algorithm is used to determine whether the chest CT scan is of a patient suffering from the COVID-19 virus or not. We used chest CT scans of COVID-19 patients obtained from~\cite{COVID19}, which is a large compilation of COVID-19 chest CT scans from several public sources~\cite{covid1, covid2, covid3, covid4, covid5, convid6}. 

The MURA dataset collected by Stanford University School of Medicine and ML Group~\cite{rajpurkar2018mura} is of X-ray scans of the patient's elbows, fingers, forearms, hands, humerus, shoulders, and wrists. This data is used to promptly diagnose those who suffer from musculoskeletal disorders. If there is a fracture present in the bone X-ray scan, then the patient is classified positive, meaning that he or she or they have this condition. On the other hand, if there is no fracture in the bones, then the patient is classified negative, as in the patient is healthy with their bones intact in one piece. 

We also analyzed real-world medical data--cholesterol levels--provided by Seoul National University Hospital (SNUH). This study was approved by the Institutional Review Board of Seoul National University Hospital (No. C-1712-009-903) with a waiver of informed consent. No personally identifiable data was included in the dataset. Data used in this study were retrieved from Seoul National University Hospital's Common Data Model (CDM) database. LDL-C is a type of cholesterol that builds up in the blood vessels and plays part in causing coronary artery disease (CAD)~\cite{hirayama2012small}. Hence, detecting the levels of LDL-C in the bloodstream is critical to prevent any health problems. The dataset obtained from SNUH contains the patient's age, sex, height, weight, total cholesterol (TC), high-density lipoprotein cholesterol (HDL-C), triglyceride (TG), and LDL-C. We train a regression model that can use the above attributes to predict the value of LDL-C as accurately as possible. 
% This study was approved by the Institutional Review Board of Seoul National University Hospital (No. C-1712-009-903) with a waiver of informed consent. No personally identifiable data was included in the dataset. Data used in this study were retrieved from Seoul National University Hospital's Common Data Model (CDM) database.

% CAN'T USE FOOTNOTES IN NATURE i checked. (Please note, footnotes should not be used.) This study was approved by the Institutional Review Board of Seoul National University Hospital (No. C-1712-009-903) with a waiver of informed consent. No personally identifiable data was included in the dataset. Data used in this study was retrieved from Seoul National University Hospital's Common Data Model (CDM) database.

% ??The number of MURA X-ray scans of 7 different body parts is shown in
% \begin{table} %[t]%\\
% \centering
% \begin{tabular}{l|ccc}
% \hline
% \centering
% Category &  Total case & Positive case & Negative case\\
% \hline
% Finger  & 5,106   & 1,968  & 3,138  \\
% Hand    & 5,543   & 1,484  & 4,059  \\
% Wrist  & 9,752   & 3,987  & 5,765 \\
% Forearm & 1,825   & 661   & 1,164 \\
% Elbow    & 4,931   &2,006  & 2,925  \\
% Humerus & 1,272   &599   & 673  \\
% Shoulder&  8,379   & 4,168  &  4,211 \\
% \hline
% \end{tabular}
% \caption{\label{tab:muradata}The number of X-ray images for each body part collected from MURA.}
% \end{table}

\subsection*{Split Learning Algorithm}
\begin{figure}[ht]
\centering
\includegraphics[width=0.97\linewidth]{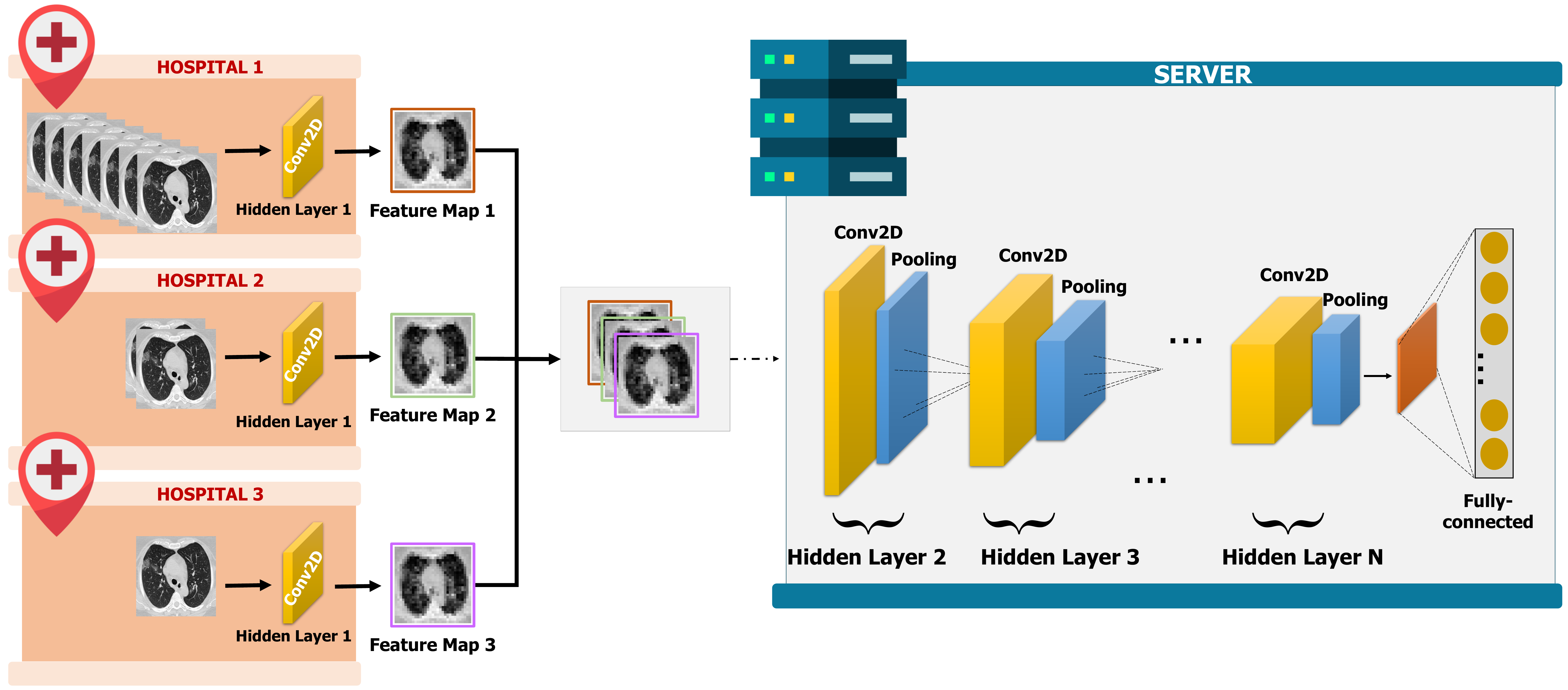}
\caption{The overall framework of our multi-site split learning algorithm.}
\label{fig:framework}
\end{figure}

In split learning, the learning process is, as the word suggests, literally split or separated into two parts: the end-systems and the server. Here, the end-systems refer to the multiple hospitals that possess original medical data to be trained in the deep learning model. The server is where the actual learning of the deep neural network occurs. Multiple hospitals are connected to one server; all the hospitals connected to the server collaborate to train the very deep neural network that is placed in the server without exposing their raw data to an external network. A meticulously devised framework of our proposed algorithm is shown in Figure  \ref{fig:framework}.  Every individual hospital only runs the deep neural network up to the first hidden layer. After passing through the first hidden layer, the end-systems send only their feature maps to the server. Since only the encrypted feature map is transferred to the external network, and not the original medical data, the privacy of confidential patient information is preserved. It is coined "multi-site split learning" since multiple hospitals (multi-site) are geometrically spaced apart from each other and the deep neural network is divided among them. 
In this study, we investigate the performance of our multi-site split learning when changes to the number of end-systems and the data distribution are made. All methods were carried out in accordance with relevant guidelines and regulations. For all experiments, the split learning setup is the same, with the deep neural network being divided amongst the end-systems and one centralized server. Even with the varying number of end-systems that are participating in training the deep neural network, each and every end-system only holds one hidden layer. This hidden layer comprises one convolutional layer. The original medical data, be it the COVID-19 CT chest scans or the MURA bone X-ray images, or the numerical cholesterol data, will pass only through the first hidden layer. The output from the first hidden layer is an encrypted feature map. This feature map is distorted to the point where it cannot be used to inference the original data in this multi-site split learning algorithm. This feature map is the only information that is exposed to an external network--the centralized server. The heavily deformed feature map that is transferred to the server is shown in Figure \ref{fig:covidhidden} and Figure \ref{fig:murahidden}. Since only this encrypted parameter is passed to the server, and not the raw medical images, protection of personal information is ensured. 

The feature maps from all the participating hospitals are concatenated and processed as the input to the deep neural network. Since the end-systems only run the neural network up to the first hidden layer, the remaining hidden layers are placed in the server. It should be noted that all the end-systems have the same number of hidden layers--that is, one hidden layer. With the concatenated feature map, the server trains the deep neural network. Considering the concatenated feature map is a summation of all the feature maps from each and every end-system, the deep neural network that is located in the server can learn using all the data from the participating hospitals. 

\subsection*{Deep Learning Models}

The deep learning models are selected appropriately to accommodate the data type used. Since the two image datasets--COVID-19 CT scans and MURA X-ray scans--require a classification tool, a classification model is used. To determine whether or not a certain patient's CT scan is that of a patient suffering from the viral COVID-19 virus, a custom classification model is designed. The epoch is set to 100, with the binary crossentropy chosen as the loss function. As for the activation function, the sigmoid function is used for this customized model. The input image sizes are all set to a size of 64 $\times$ 64 $\times$ 1, with a batch size set to 64. This setting is applied to the server where it trains to classify the CT scan from that of a COVID-19 patient to a healthy lung. The end-systems hold one hidden layer while the server trains with 4 hidden layers. 

As for the MURA datasets, a deeper VGG19 model is used to train the classification model. Due to the fact that there is a lower number of data compared to the 407,953 images of COVID-19 CT scans, a deeper and more intricate neural network is used. This model is devised with an epoch of 50 and a batch size of 128. The loss and activation functions are binary crossentropy and sigmoid, respectively. The X-ray image sizes are all reshaped to input size of 224 $\times$ 224 $\times$ 1. The MURA bone X-rays scans are trained with a total of 20 layers: one hidden layer placed in each hospital and the remaining 19 layers placed in the server, where the actual training of VGG19 occurs. 

On the other hand, cholesterol data is a numerical dataset. Hence, a prediction model must be used. The deep learning model configuration for each of the three datasets is summarized in Table \ref{tab:dlsetup}. To accurately predict the level of LDL-C in the bloodstream, we designed a custom regression model with an epoch of 200, loss function of mean squared error (MSE), the activation function of Leaky ReLU, and a batch size of 2,048. The input size for this numerical data represents the number of data used to train the custom regression model. This prediction model has 3 layers, of which one is placed in the contributing hospitals, and the remaining 2 is placed in the server.
\begin{table}[ht]
\centering
\begin{tabular}{l|ccc}
\hline
\centering
Parameters & COVID-19& MURA & Cholesterol\\
\hline
Epochs  & 100 & 50 & 200 \\
Loss & Binary crossentropy & Binary crossentropy & MSE  \\
Activation function & Sigmoid & Sigmoid & Leaky ReLU \\
Batch size & 64 & 128 & 2,048 \\
Input Size & 64 $\times$ 64 $\times$ 1 & 224 $\times$ 224 $\times$ 1  & 326,032  \\
Model & Custom classification & VGG19 & Custom regression  \\
\hline
\end{tabular}
\caption{\label{tab:dlsetup}The arrangements for a deep learning model to train each of the COVID-19, MURA and cholesterol data sets.}
\end{table}

A generalization of the deep neural network configuration is shown on the server's side in Figure \ref{fig:framework}. A hidden layer is a group of a Conv2D layer and MaxPooling2D layer. This framework is a generalization of the deep learning structure that is applied to the three datasets. As it can be concluded, our multi-site split learning algorithm applies to all kinds of data types if the adequate model is selected. 

\section*{Results}

\subsection*{Preserving Privacy of Medical Data}
The anticipation of today's rapid epochal shift into the information and technology era extends into the interoperability of medical data with external research partnerships. The sharing of medical information between health care provides is extremely advantageous to both the patient and physicians as well. The global health crisis is shining a harsh spotlight on healthcare’s biggest loopholes; the benefits of data sharing come hand-in-hand with the risks of unscrupulous hackers getting access to personal information. The ominous presence of hackers in cyberspace creates a fear of leaks of sensitive patient information wherever it gets transferred. 

Therefore, one of the paramount benefits of our proposed multi-site split learning algorithm is the protection of the original training data. Since the deep neural networks are, as the word suggests, literally split between the multiple end-systems and the server. Each of the participating hospitals, or end-systems, only has one hidden layer, and the rest of the hidden layers are located in the server. After running the deep neural network up to the first hidden layer, the feature maps from all the active hospitals are sent to the server, where they are concatenated. This concatenated feature map becomes the input to the server's deep neural network. The feature map is the sole information that is exposed to a third-party affiliation; the original medical data does not get shared between the participating hospitals.

The extent to which the raw medical data is protected is shown in the following Figures. Figure \ref{fig:covidhidden} shows the degree to which the original COVID-19 CT scan is distorted after passing through one hidden layer of each hospital. Figure \ref{fig:covidhidden} (a) is the original CT scan of a COVID-19 patient. Figure \ref{fig:covidhidden} (b) is the highly deformed image of image (a) after it undergoes the first hidden layer. Each hospital will produce an image like Figure \ref{fig:covidhidden} (b). All of these feature maps are concatenated and communicated to the server where it then trains the deep neural network. 

\begin{figure}[H]
\centering
\setlength{\tabcolsep}{3pt}
\renewcommand{\arraystretch}{0.25}
\begin{tabular}{cc}
\includegraphics[width=0.35\linewidth]{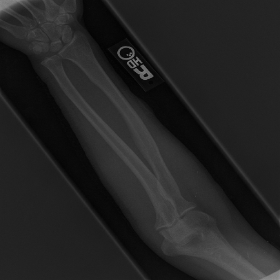} &
\includegraphics[width=0.35\linewidth]{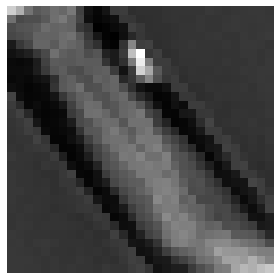}
\tabularnewline
(a) & (b)
\end{tabular}
\caption{Ensuring privacy of original medical data. Image (a) is the original image of the MURA bone X-ray scans. Image (b) shows the highly distorted image after passing through one hidden layer at the hospitals.}
\label{fig:murahidden}
\end{figure}

Similarly, the protection of patient records applies to MURA bone X-ray scans as well. Each hospital possesses raw images of MURA bone X-ray scans. These hospitals run the deep neural network only up to the first hidden layer using their own raw medial data, which is displayed in Figure \ref{fig:murahidden} (a). The output of the first hidden layer, which is depicted in Figure \ref{fig:murahidden} (b), is collected from each of the end-systems. The second image is the only information that is revealed to the server. Hence, the original medical image is protected from any and all external threats. Furthermore, as evidently demonstrated in the comparison between Figure \ref{fig:murahidden} (a) and (b), even if hackers do obtain possession of the feature maps when transferring to the server, it will be very difficult to trace it back to the original image. Once again, protecting the privacy of patient information. 

\begin{figure}[ht]
\centering
\setlength{\tabcolsep}{3pt}
\renewcommand{\arraystretch}{0.25}
\begin{tabular}{cc}
\includegraphics[width=0.35\linewidth]{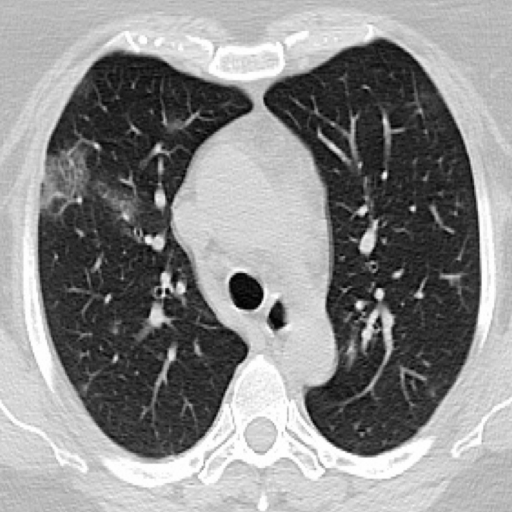} &
\includegraphics[width=0.35\linewidth]{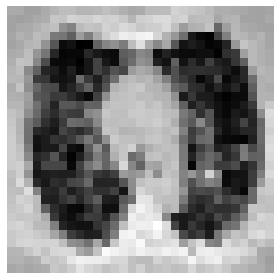}
\tabularnewline
\tabularnewline
(a)    & (b) 
\end{tabular}
\caption{Ensuring privacy of original medical data. Image (a) is the original image of the COVID-19 CT scan. Image (b) shows the highly distorted image after passing through one hidden layer at the hospitals.}
\label{fig:covidhidden}
\end{figure}

The same principle applies to numerical data. Just as the combination of the Conv2D and MaxPooling2D layer, in the server's end shown in Figure \ref{fig:framework}, alters the original data to a state where it is unrecognizable, the original cholesterol data are also deformed to protect the raw patient records in the same manner.

% \subsubsection*{Third-level section}
 
% Topical subheadings are allowed.

\subsection*{Effect of Varying Number of End-systems}

It is conspicuous that data exchange regarding medical information opens doors for meaningful research. Our split learning model is a cohesive ecosystem that seamlessly works to achieve one goal, which is to provide quality care for patients. It is a safe and trustworthy technique that entities outside of a particular hospital, such as pharmaceutical companies, clinical and molecular research centers, can leverage their expertise to avoid medication errors. The pandemic has instilled the need for patient-centric data sharing across clinical institutions to help understand the integral effect of this unprecedented virus. Extending further from just the pandemic, sharing medical data among other hospitals can reduce doctor visits or hospital admissions. 

In this experiment, we examine the effect of differing the number of end-systems on classification performance. Our algorithm is set up to emulate a real case scenario where multiple hospitals come together to share their sources and collaborate to train a deep neural network. As discussed in the introduction, more hospitals are wanting to co-operate with others in medical research and data-sharing. We test the outcomes according to the number of participating hospitals, accurately presenting when the split learning can achieve better performance by collaboration.

Hence, the experiment is devised with a varying number of end-systems, i.e., hospitals. Each of the hospitals transfers only the feature maps up to the centralized server. In our split learning design, the deep neural network is divided in the following way. Each hospital runs the neural network up to the first hidden layer using its own unique data. Then the output of that first hidden layer, which is the feature map, is sent to the server. The server takes the feature map as input and continues training with the rest of the hidden layers. Hence, the end-systems only have one hidden layer, and the server is comprised of the rest of the hidden layers, which is where the majority of the computation is performed. This divided deep neural network setup is the same for all the participating hospitals, whether there are 3, 4, or 5 contributing hospitals. 

% , to test the effect of increasing the number of participating hospitals

\subsubsection*{Result on COVID-19}
This experiment classifies the CT scans into two groups: those suffering from SARA-CoV-2 and those healthy without any chest pain. The chest CT scans of COVID-19 patients and non-COVID-19 patients are processed under a convolutional neural network (CNN). Using a loss function of binary crossentropy, an activation function of sigmoid, a batch size of 64, and an input size of 64 $\times$ 64 $\times$ 1, the CT scan images are classified. The experiment setup is summarized in Table~\ref{tab:dlsetup}.

In this work, we test the efficacy of increasing the number of cooperating hospitals in a split learning environment. Furthermore, we give an empirical guideline to the optimal number of hospitals for the highest accuracy with the lowest loss. The two major patterns of increasing the number of end-systems only are highlighted below: 
\begin{itemize}
\item Overfitting can be prevented when there are fewer end-systems, yet it is difficult for performance to improve as the number of end-systems increases. 
% The classification accuracy decreases as the number of end-systems increases.
\item Once the number of end-systems reaches five, loss takes longer to converge and stabilize. 
% \item As the split ratio changes to incorporate one end-system, which possess majority of the data, classification accuracy improves. 
\end{itemize}

\begin{figure*}[h!]
\centering
\setlength{\tabcolsep}{3pt}
\renewcommand{\arraystretch}{0.25}
\begin{tabular}{cc}
\includegraphics[page=1, width=0.5\textwidth]{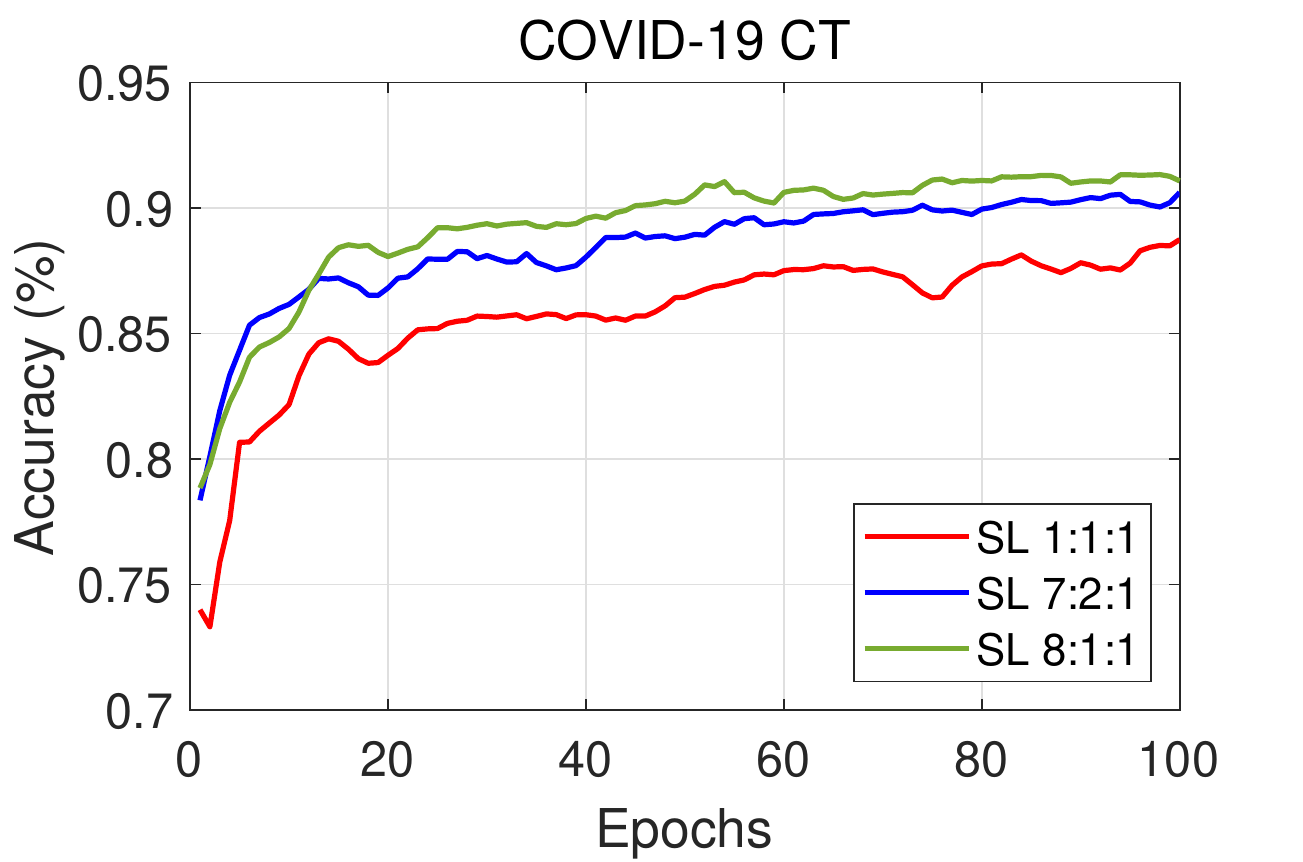} &
\includegraphics[page=1, width=0.5\textwidth]{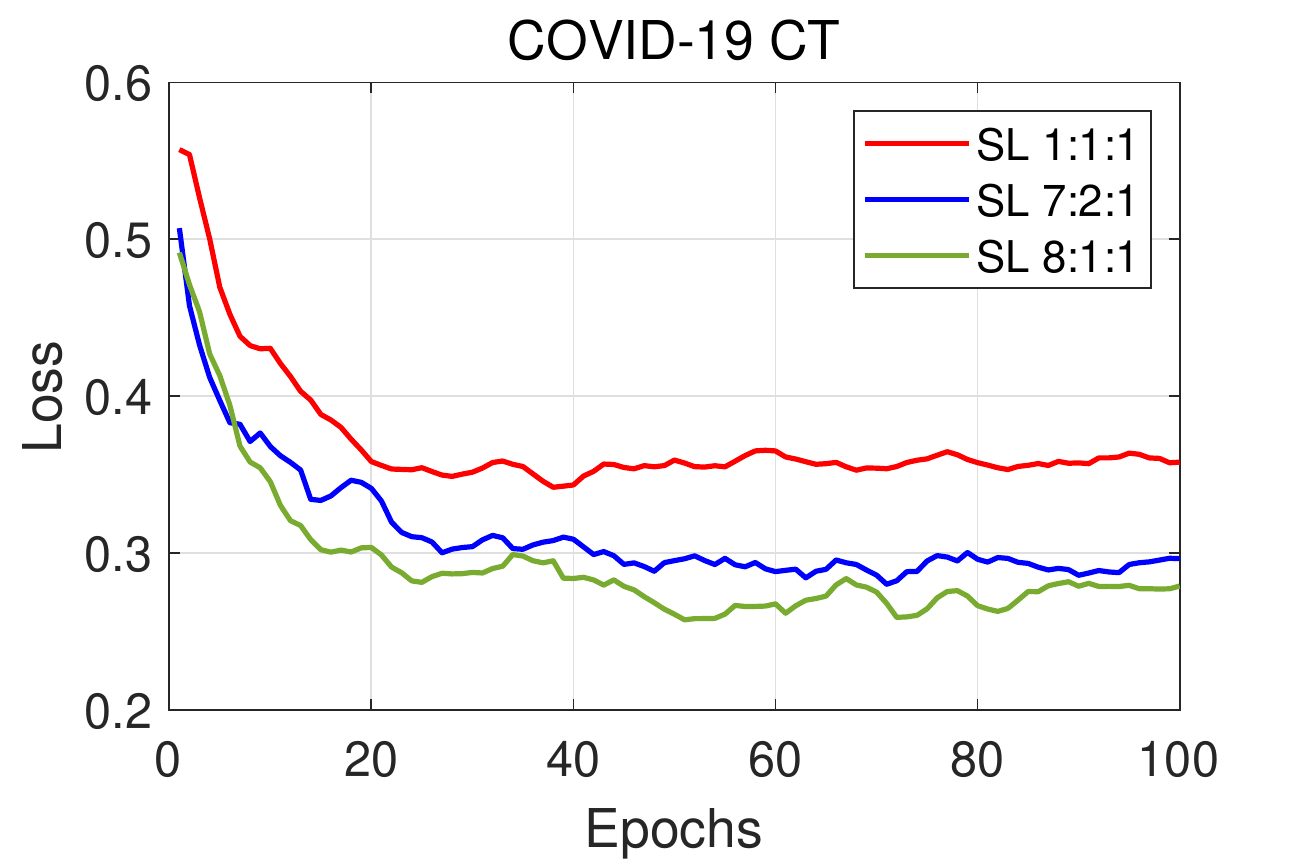} 
\tabularnewline
\tabularnewline
(a)  Accuracy with 3 end-systems. & (b) Loss with 3 end-systems. 
\end{tabular}
\caption{Performance tested with three end-systems. The graph depicts the classification results with three end-systems having 1:1:1, 7:2:1 and 8:1:1 data-imbalance.}
\label{fig:n3}
\end{figure*}

\begin{figure*}[h!]
\centering
\setlength{\tabcolsep}{3pt}
\renewcommand{\arraystretch}{0.25}
\begin{tabular}{cc}
\includegraphics[page=1, width=0.5\textwidth]{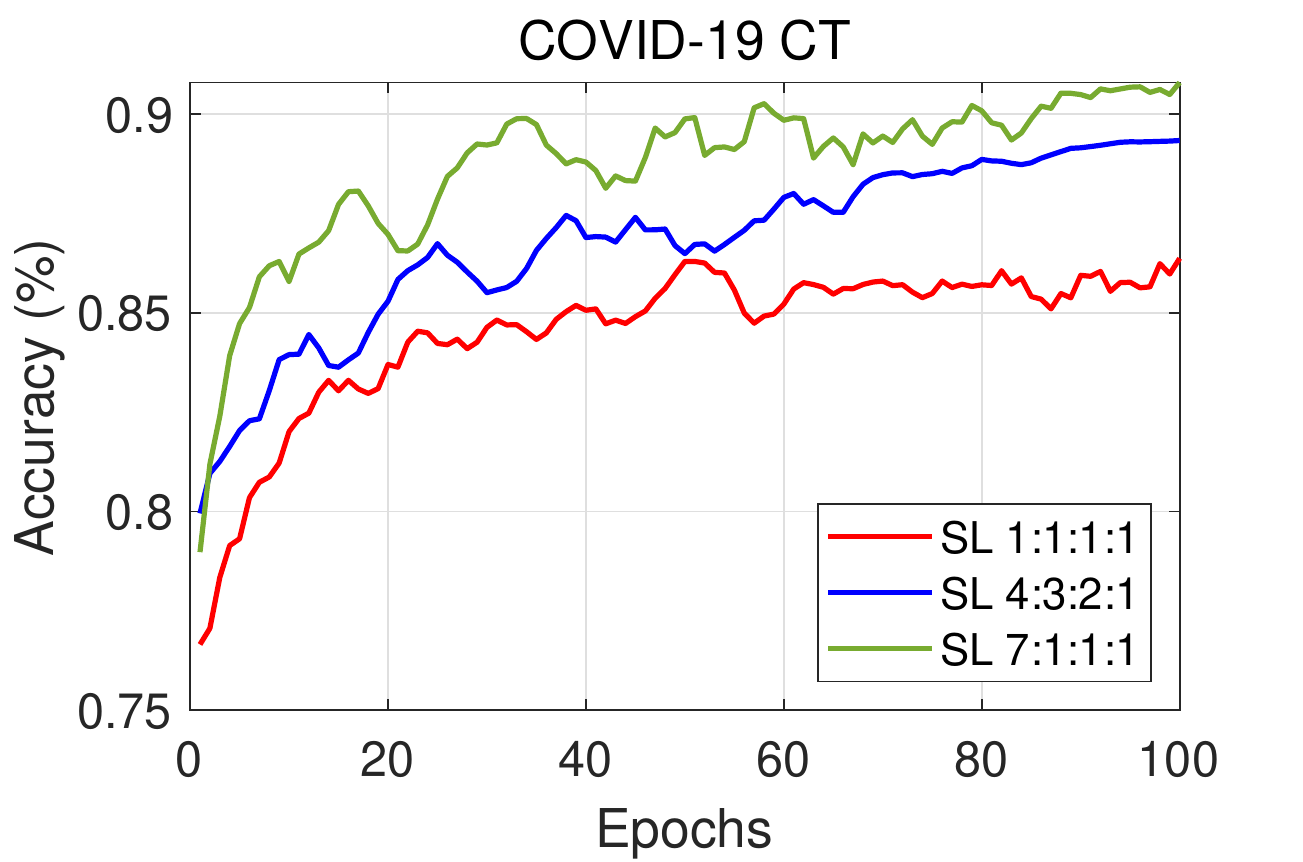} &
\includegraphics[page=1, width=0.5\textwidth]{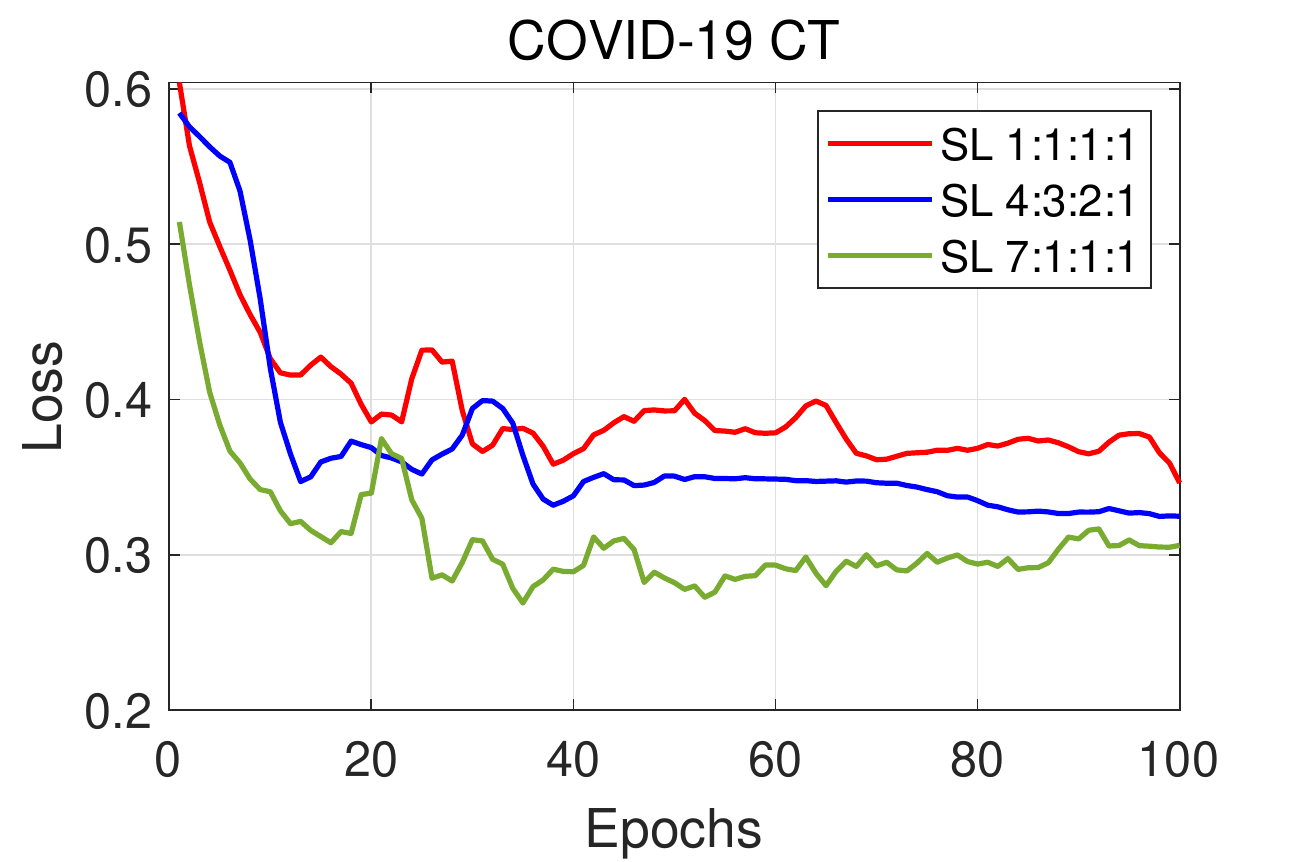} 
\tabularnewline
\tabularnewline
(a)  Accuracy with 4 end-systems. & (b) Loss with 4 end-systems.
\end{tabular}
\caption{Performance tested with four end-systems. The graph depicts the classification results with four end-systems having 1:1:1:1, 4:3:2:1 and 7:1:1:1 data-imbalance.}
\label{fig:n4}
\end{figure*}

\begin{figure*}[h!]
\centering
\setlength{\tabcolsep}{3pt}
\renewcommand{\arraystretch}{0.25}
\begin{tabular}{cc}
\includegraphics[page=1, width=0.5\textwidth]{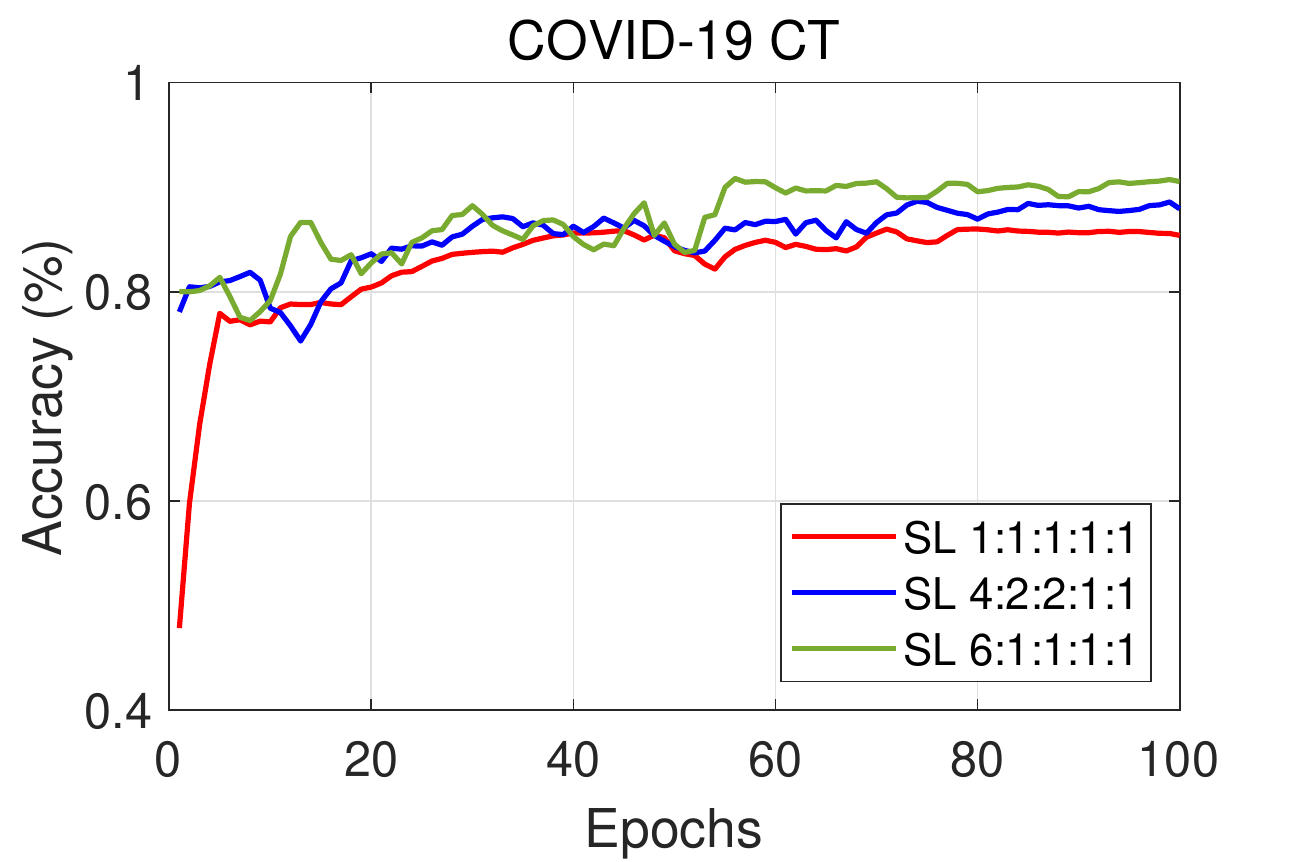} &
\includegraphics[page=1, width=0.5\textwidth]{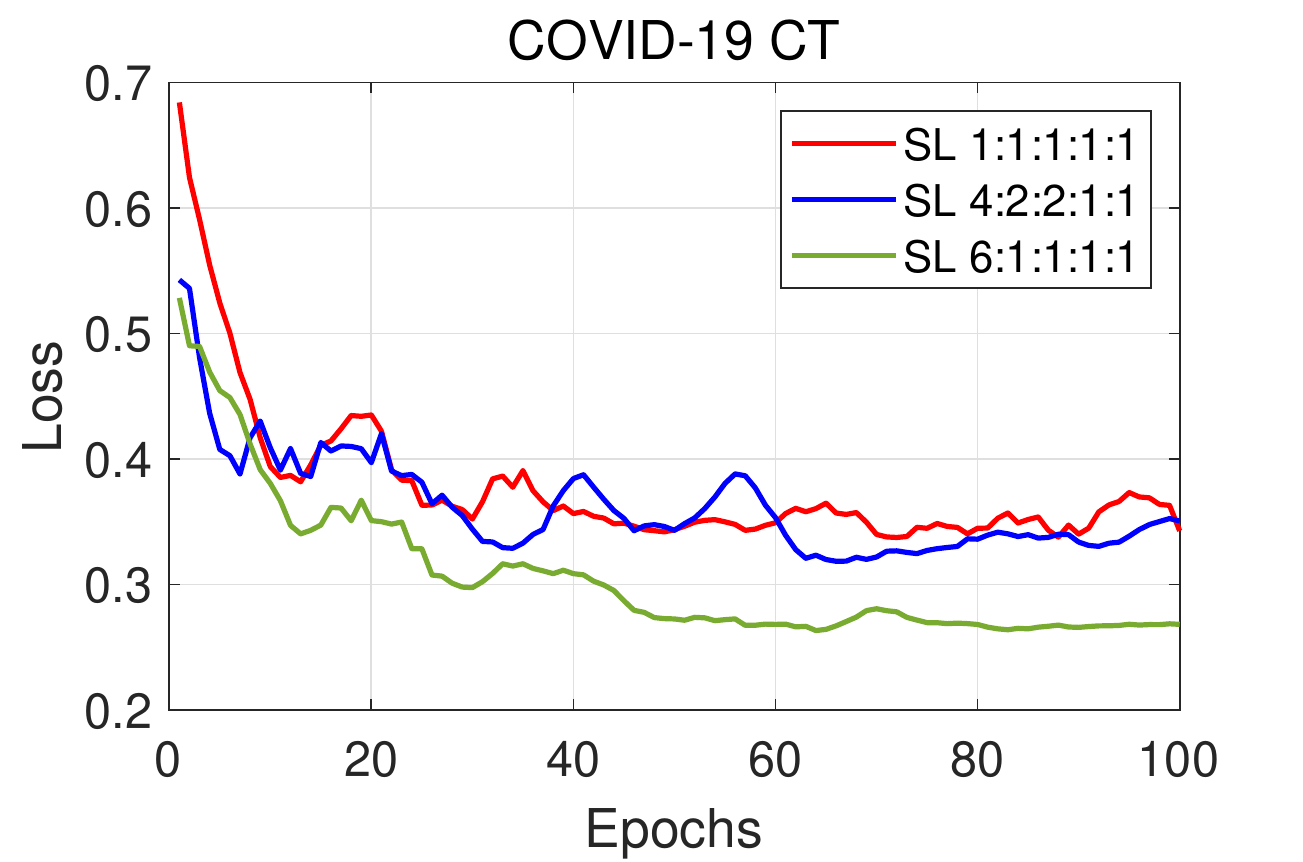} 
\tabularnewline
\tabularnewline
(a)  Accuracy with 5 end-systems. & (b) Loss with 5 end-systems.
\end{tabular}
\caption{Performance tested with five end-systems. The graph depicts the classification results with four end-systems having 1:1:1:1:1, 4:2:2:1:1 and 6:1:1:1:1 data-imbalance.}
\label{fig:n5}
\end{figure*}

Comparing Figure~\ref{fig:n3} with Figure~\ref{fig:n4} and with Figure~\ref{fig:n5}, a visual contrast between the accuracy and loss patterns can be made. Only considering the increasing number of end-systems, and not the split ratios, it is evident that the classification accuracy decreases as more hospitals join split learning. This pattern can be seen when comparing the subsequent data ratio of each end-system with its corresponding ratio.
An equal division of the datasets between the corresponding end-systems is a default ratio and will be the acting benchmark for fair comparisons. Otherwise stated, an equally divided 1:1:1 ratio, a 1:1:1:1 ratio, and a 1:1:1:1:1 ratio for when there are 3, 4, and 5 end-systems respectively. This marks the base for a fair comparison rather than looking at the 7:2:1, 4:3:2:1, and 4:2:2:1:1 ratio group to make conclusions on varying the number of end-systems.

% side by side gives a rather equal comparison for the
% Similarly, 8:1:1, 7:1:1:1, and 6:1:1:1:1 are comparable when judging the effect of varying the number of end-systems, which will be discussed in the next section 

% that there is no distinct difference in the classification accuracy between 3 and 4 end-systems. However, once the number of hospitals participating in split learning reaches five, the accuracy in classifying whether a patient is suffering from COVID-19 or not drops. 

The same observation can be made with proof from the loss graphs. It is more meaningful to judge the rate of convergence in the case of loss graphs. In comparing Figure~\ref{fig:n3} (b) and Figure~\ref{fig:n4} (b), it is evident that a lower loss value is reached at a smaller amount of epochs for 3 end-systems than that of 4 end-systems. The loss curves for 3 hospitals stabilize quicker and converge at a relatively lower loss value. When one more end-system is added and Figure~\ref{fig:n5} (b) comes into the picture, the loss value reaches a stable point at a higher level. This validates our claim that the performance levels fall as more hospitals engage in split learning. 
% the loss values don't differ drastically to each other. 

\begin{table}[H]
\begin{tabular}{|c|c|c|c|c|c|c|c|c|c|}
\hline
Number of end-systems & \multicolumn{3}{c|}{3} & \multicolumn{3}{c|}{4}      & \multicolumn{3}{c|}{5}            \\ \hline
Split ratio           & 1:1:1  & 7:2:1 & 8:1:1 & 1:1:1:1 & 4:3:2:1 & 7:1:1:1 & 1:1:1:1:1 & 4:2:2:1:1 & 6:1:1:1:1 \\ \hline
Accuracy(\%)          & 88.3  & 90.1 & 91.4 & 85.7  & 89.3  & 90.8   & 85.6   & 88.8    &  90.5 \\ \hline
\end{tabular}
\caption{\label{tab:coco}Classification accuracy using COVID-19 CT scan.}
\end{table}

Furthermore, Table~\ref{tab:coco} shows the exact accuracy level at which all the end-systems converge. Having these values displayed side-by-side, the highest accuracy level of 91.4\% is achieved by our split learning algorithm experimented with three end-systems. 
% Judging solely on the number of end-systems done with our experimental line-up, it can be empirically induced that the best option for split learning to achieve optimal performance is: five end-systems with a 6:1:1:1:1 ratio distribution.

\subsubsection*{Result on MURA}
The same test is conducted using MURA datasets and the experiment setup is summarized in Table~\ref{tab:dlsetup}. Here, we want to observe the effect of increasing the number of end-systems in split learning has on the classification accuracy using a different medical dataset. Since this test is conducted with MURA, it gets X-ray scans of each of the seven body parts as inputs:fingers, elbows, forearms, hands, humerus, shoulders, or writs. The X-ray scans go through one hidden layer at the hospital and the output feature map is sent to the server. Once the feature map is received at the server's end, the concrete CNN procedure proceeds. The server will determine whether or not the patient is suffering from musculoskeletal disorders from the feature maps of the X-ray scans. 

\begin{figure}[h!]
  \centering
  \begin{subfigure}{.35\linewidth}
    \centering
    \includegraphics[width = \linewidth]{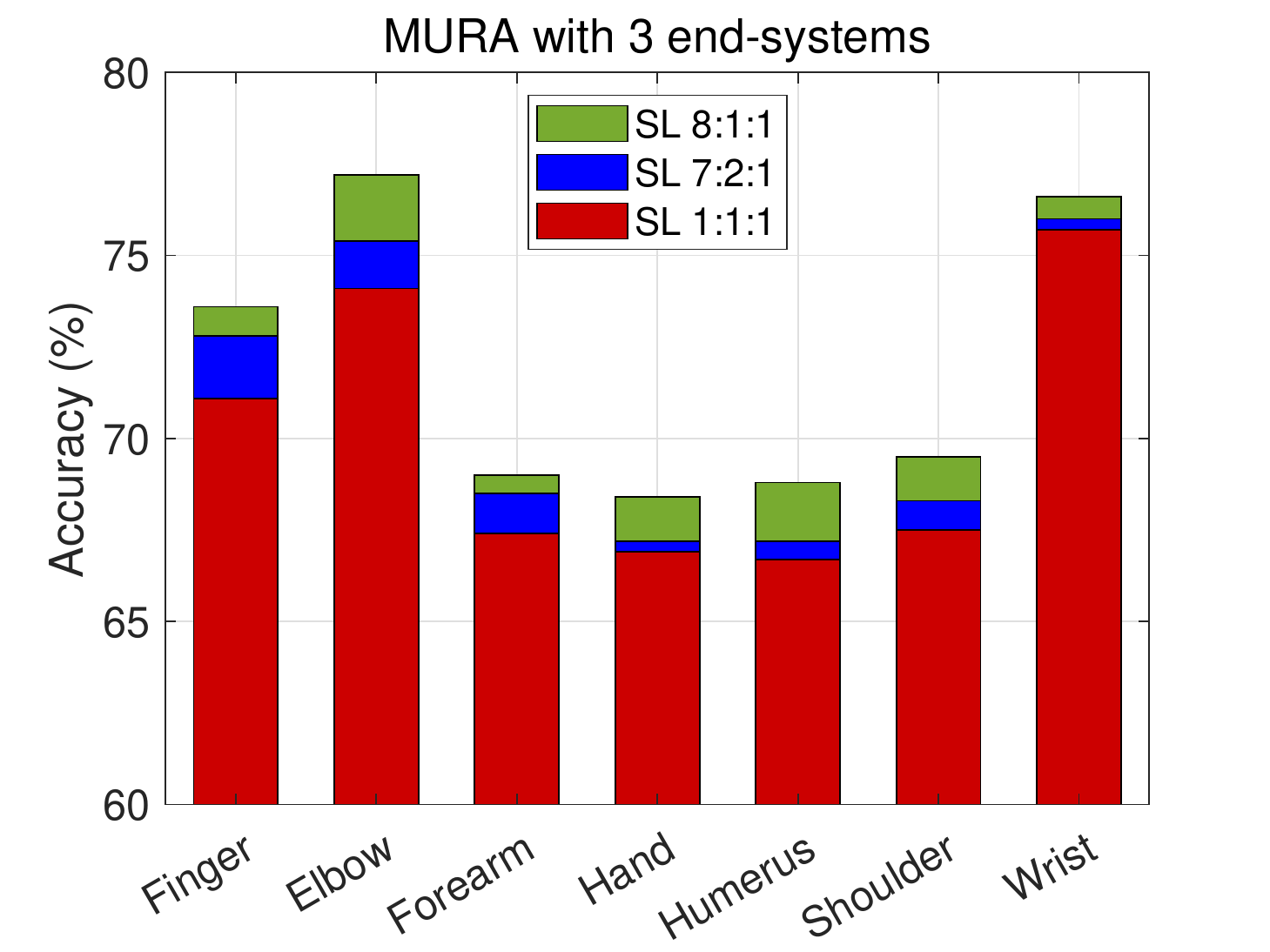}
    \caption{Accuracy with 3 end-systems.}
  \end{subfigure}%
  \begin{subfigure}{.35\linewidth}
    \centering
    \includegraphics[width = \linewidth]{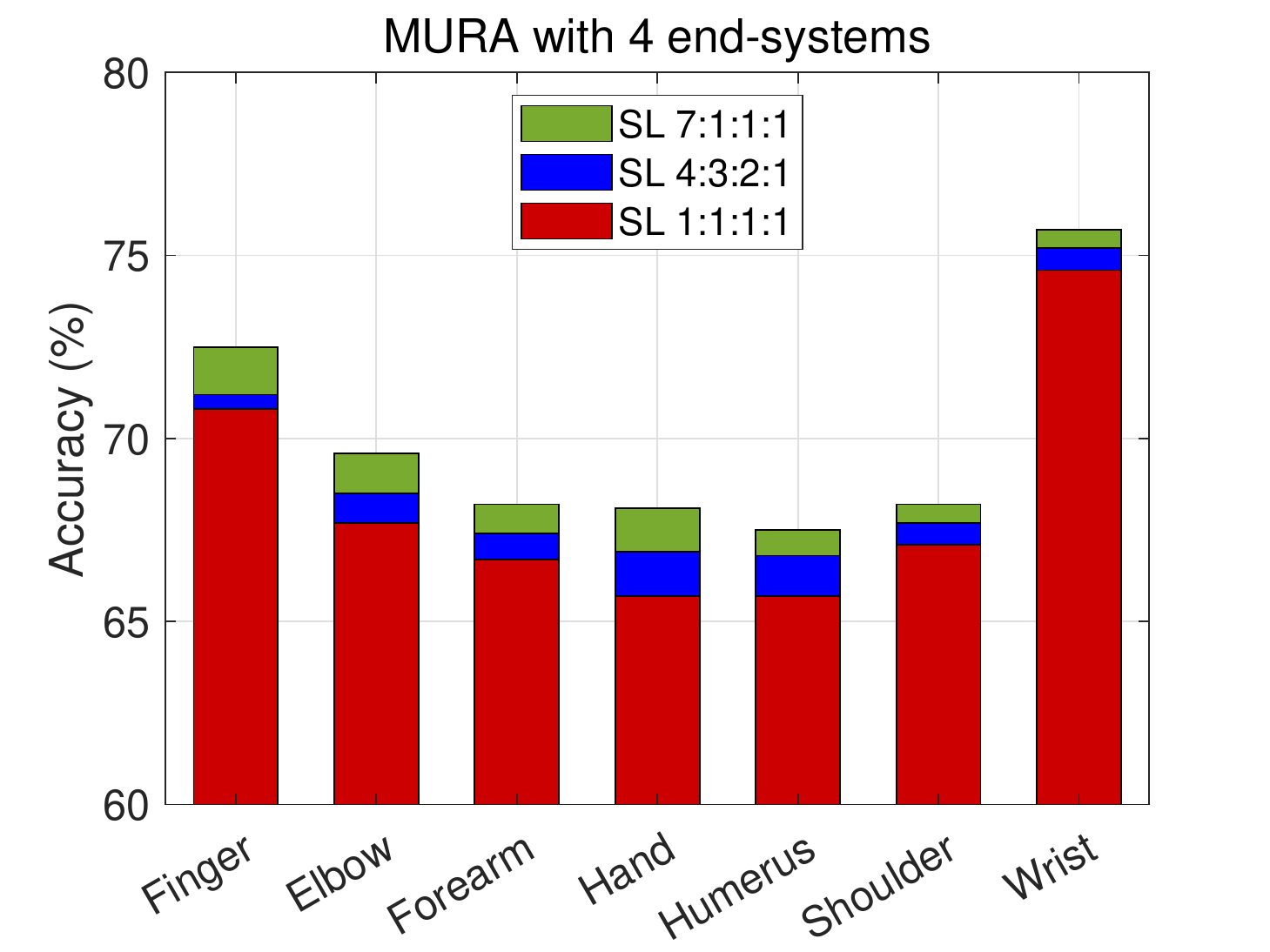}
    \caption{Accuracy with 4 end-systems.}
  \end{subfigure}%
  \begin{subfigure}{.35\linewidth}
    \centering
    \includegraphics[width = \linewidth]{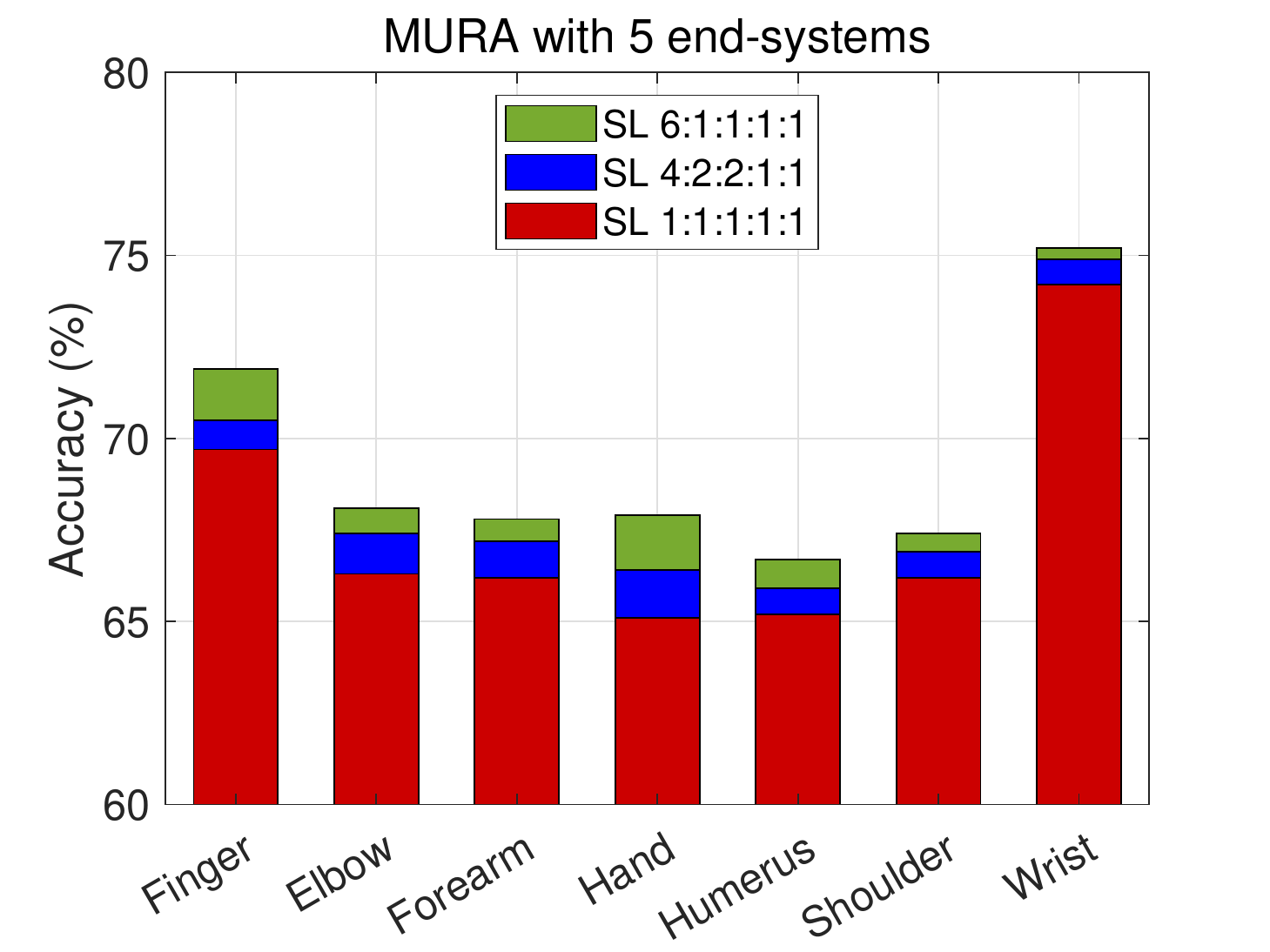}
    \caption{Accuracy with 5 end-systems.}
  \end{subfigure}
  \caption{\label{fig:muraacc}Accuracy of the seven body parts using MURA dataset for (a) 3, (b) 4, and (c) five end-systems.}
\end{figure}

\begin{table}[h!]
\begin{tabular}{lllllllllll}
\hline
\multicolumn{11}{c}{\textbf{MURA Accuracy(\%)}}                                                                                           \\ \cline{2-11}  & \multicolumn{1}{l|}{\textbf{\begin{tabular}[c]{@{}l@{}}Number of \\ end-systems\end{tabular}}} & \multicolumn{3}{c|}{3}                     & \multicolumn{3}{c|}{4}                           & \multicolumn{3}{c}{5}             
\\ \cline{2-11}  & \multicolumn{1}{l|}{\textbf{Split ratio}} & 1:1:1 & 7:2:1 & \multicolumn{1}{l|}{8:1:1} & 1:1:1:1 & 4:3:2:1 & \multicolumn{1}{l|}{7:1:1:1} & 1:1:1:1:1 & 4:2:2:1:1 & 6:1:1:1:1 \\ \hline
\multirow{7}{*}{\textbf{\begin{tabular}[c]{@{}l@{}}MURA \\ body \\ parts\end{tabular}}} & \multicolumn{1}{l|}{Finger} & 71.1  & 72.8  & \multicolumn{1}{l|}{73.6}  & 70.8    & 71.2    & \multicolumn{1}{l|}{72.5}    & 69.7      & 70.5      & 71.9      \\ & \multicolumn{1}{l|}{Elbow} & 74.1  & 75.4  & \multicolumn{1}{l|}{77.2}  & 72.9    & 73.2    & \multicolumn{1}{l|}{75.2}    & 72.0      & 72.6      & 73.7      \\  & \multicolumn{1}{l|}{Forearm}   & 67.4  & 68.5  & \multicolumn{1}{l|}{69.0}  & 66.7    & 67.4    & \multicolumn{1}{l|}{68.2}    & 66.2      & 67.2      & 67.8      \\ & \multicolumn{1}{l|}{Hand}   & 66.9  & 67.2  & \multicolumn{1}{l|}{68.4}  & 65.7    & 66.9    & \multicolumn{1}{l|}{68.1}    & 65.2      & 66.4      & 67.9      \\& \multicolumn{1}{l|}{Humerus}  & 66.7  & 67.2  & \multicolumn{1}{l|}{68.8}  & 65.7    & 66.8    & \multicolumn{1}{l|}{67.5}    & 65.2      & 65.9      & 66.7      \\ & \multicolumn{1}{l|}{Shoulder}  & 67.5  & 68.3  & \multicolumn{1}{l|}{69.5}  & 67.1    & 67.7    & \multicolumn{1}{l|}{68.2}    & 66.2      & 66.9      & 67.4      \\ & \multicolumn{1}{l|}{Wrist} & 75.7  & 76.0  & \multicolumn{1}{l|}{76.6}  & 74.6    & 75.2    & \multicolumn{1}{l|}{75.7}    & 74.2      & 74.9      & 75.2      \\ \hline
\end{tabular}
\caption{\label{tab:muraacc}Classification accuracy using MURA body parts.}
\end{table}

The results for this experiment are graphed in Figure~\ref{fig:muraacc}, with the classification accuracy for each body part shown. In comparison between the number of end-systems, we look at the highest accuracy out of the three split ratios. With the finger X-ray images, the performance with 3 hospitals training CNN under equally divided data ratio has a classification accuracy of 71.1\%, with 4 it drops to an accuracy of 70.8\%, and with 5 end-systems it achieves \typeout{96.7} 69.7\%, as seen in Table~\ref{tab:muraacc}. For the elbow scans, split learning with 3 end-systems reaches up to 74.1\% accuracy, 72.9\% for 4 end-systems, and 72.0\% for 5 hospitals. Analyzing the rest of the body parts, a general trending pattern is observed. When only considering the number of end-systems, it is concluded that 3 end-systems give the best image classification performance\typeout{ and}. Therefore, as the number of end-systems increases, the classification accuracy level decreases. This result coincides with the analysis made with the COVID-19 chest CT scan data as well.

% like so, we can generalize a trending pattern. 
% There are four cases in elbows, the humerus, shoulders and wrists that have the highest classification accuracy when trained with only 3 end-systems in the split learning procedure. The finger, forearm, and hand dataset have the highest accuracy when trained with 4 end-systems.

\subsubsection*{Result on Cholesterol}
The split learning algorithm is a versatile tool that can train any type of data. With an appropriate model selected, split learning can be used to train image data or even numerical data as well. With the previous two COVID-19 and MURA scans, we saw that image data is processed to give aid in medical treatment plans. In this section, we see the results of split learning using cholesterol data. The experiment is conducted to predict the so-called Low-density lipoprotein cholesterol (LDL-C) level using other patient attributes such as age, sex, height, and weight. This regression model is used to predict the level of LDL-C in the blood and the performance is tested using root mean squared logarithmic error (RMSLE). RMSLE is a standard means to compare the loss value between the original and the predicted data in a regression model.

\begin{table}[h!]
\begin{tabular}{|c|c|c|c|c|c|c|c|c|c|}
\hline
Number of end-systems & \multicolumn{3}{c|}{3}   & \multicolumn{3}{c|}{4}      & \multicolumn{3}{c|}{5}            \\ \hline
Split ratio       & 1:1:1  & 7:2:1  & 8:1:1  & 1:1:1:1 & 4:3:2:1 & 7:1:1:1 & 1:1:1:1:1 & 4:2:2:1:1 & 6:1:1:1:1 \\ \hline
RMSLE             & 0.0613 & 0.0597 & 0.0567 & 0.0651  & 0.0624  & 0.0593  & 0.0674    & 0.0658    & 0.0616    \\ \hline
\end{tabular}
\caption{\label{tab:chol}The RMSLE loss values for cholesterol data. It shows the loss for each end-system (3, 4, 5 end-systems) with varying data-ratios.}
\end{table}

Looking at the results of the experiment tested with varying the number of end-systems in Table~\ref{tab:chol}, a noticeable and repeating pattern can be observed. The table shows the loss value, i.e. the difference between the actual and predicted value, the lower the loss, the better the result. RMSLE records a loss of 0.0613 with 3 end-systems, 0.0651 with 4 end-systems, and 0.0674 loss with 5 end-systems when the data are equally split. Hence, a general trend is noted with the cholesterol data--with an increasing number of end-systems, the performance of split learning falls. This observation is consistent with the trends seen using the two image datasets. Therefore, when the number of end-systems is considered only, the highest prediction performance is seen when 3 hospitals engage in split learning.

\subsection*{Effect of Changing the Data Ratio}
Previously in the above section, this paper explored the impact of varying the number of end-systems, i.e. hospitals, has on the performance of our split learning algorithm. In this section, we will investigate the outcome of changing the data-imbalance split ratio. Data-imbalance is a prevalent issue that occurs in data-sharing. This experiment is strategically calculated to simulate the real-life issue medical sectors face. In reality, one hospital may possess a large number of data on magnetic resonance imaging (MRI). Yet, another hospital may lack data in that department, which makes that hospital inadequate in making an accurate informed medical diagnosis. This is where our proposed split learning comes in handy. When this small hospital becomes a participant in this split learning algorithm, it can seek the benefits of other hospital's data without risk of exposing sensitive patient information. In our study, the data ratio is a representation of the hospital scale--a small split ratio portion is assigned to an end-system that holds a small amount of data, which is symbolic of a small-scale hospital. Similarly, the portion that is a relatively big data split ratio represents a large-scale hospital that holds voluminous amounts of data. 

Therefore, to reflect on this reality, the experiment is arranged so that each of the training datasets is divided into different ratios. For example, if 4 hospitals are collaborating to build a deep neural network, we can divide the datasets into a 4:3:2:1 ratio. That is, one hospital is assigned to have 40\% of the data, the other hospital will possess 30\% of the data, another hospital will have 20\% of the data, and the last hospital will have the remaining 10\% of the data. By dividing the datasets into these unique ratios, our study speculates the reality of data-imbalance in the medical sector and delves into exploring the effect of having different amounts of data in split learning. 

Referring back to Table~\ref{tab:coco}, a deep analysis on the effect of varying the split ratio is conducted. Three different ratio divisions are considered for each of the separate end-systems. First of all, let's have a look at the split ratio for when 3 hospitals are training a deep neural network. When all 3 end-systems divide the COVID-19 dataset equally, the classification accuracy is 88.3\%. As it is rather unlikely that all 3 hospitals share an equal amount of data, the data is split into a 7:2:1 ratio among the 3 end-systems. This gives a classification accuracy for COVID-19 chest CT scans of 90.1\%. Then, another variation is given, with a ratio of 8:1:1, which gives 91.4\% performance. For the same reasoning, a split among the datasets is made when 4 hospitals engage in split learning. As seen in Table~\ref{tab:coco} and Figure~\ref{fig:n4}, with a 1:1:1:1 ratio, the accuracy level reaches 85.7\%. A split ratio of 4:3:2:1 performs 89.3\% and that of 7:1:1:1 reaches a classification accuracy of 90.8\%. Similarly, test results for the data split when there are 5 end-systems are shown in Table~\ref{tab:coco} and Figure~\ref{fig:n5}. The experimental results show that as the split ratio varies from equal (1:1:1:1) to data-imbalance (4:3:2:1) to extreme data-imbalance (7:1:1:1), the classification accuracy increases for all 3, 4, and 5 end-systems. 

% A 1:1:1:1:1 ratio gives an accuracy level of 85.6\%, a split ratio of 4:2:2:1:1 attains a 88.8\% accuracy, and a 6:1:1:1:1 ratio fulfills up to 90.5\% in performance. 

The same experiment is conducted in the same manner using MURA bone X-ray data. The results of all the individual split ratios with a different number of end-systems for every seven body parts are represented in Figure ~\ref{fig:muraacc}. The exact accuracy levels are summarized in Table~\ref{tab:muraacc}. For all the seven body parts, the split ratio that comprises of a large data portion along with equally split data ratios (i.e., 8:1:1, 7:1:1:1, and 6:1:1:1:1) executes the best performance for the three different end-systems. The results show a clear trend in that the performance is at its peak when the data ratio is split equally with one end-system possessing a majority of the data. In other words, with 5 participating hospitals, the highest accuracy level is achieved when the data is split into a ratio of 6:1:1:1:1, where one hospital has a large portion of the data and the remaining end-systems have an equal amount of data.

Similarly, the effect of varying the split ratio is tested using cholesterol data as well. As it can be observed from the test results in Table~\ref{tab:chol}, the best performances are given when there a group of smaller-sized hospitals joins forces with one big scale hospital. That is to say when the split ratio is distributed so that one end-system holds a large portion of the data and the other end-systems have equally shared the remaining data, the RMSLE loss value is the lowest, hence accomplishing the best outcome. Consequently, the generic pattern monitored across all three different datasets for variations made to the split ratio is summarized as: 
\begin{itemize}
    \item The highest performance in split learning is observed when a hospital holding a large amount of data is paired with other small hospitals that each possess an equal amount of data. 
\end{itemize}

% is an absence in the data-imbalance problem.% is equally divided amongst all the participating end-systems, 

% describe in detail what i see in the graph when the data ratio changes for each number of clients. so i will have three sections.

% I might list the main effects of changing the data ratio to make it easier for readers to view. 
% \begin{itemize}
% \item First item
% \item Second item
% \end{itemize}

\section*{Discussion}
% The Discussion should be succinct and must not contain subheadings.

% - talk about the combined effect of end system and split ratio. 
% - concluding final pattern we see. give an empirical conclusion on the best ratio and best client number. 
% - explain why the results are the way they are. for changing the number of clients and the data ratio. 

% {\color{red}emphasize that the the minisicule decrease in accuracy for when the number of hospitals rise is able to be overlooked, ---> because it allows more hosptials to participate.
% - FORM MORE PARTNERSHIPS FOR VALUABLE MEDICAL BREAK THROUGHS
% - e.g with COVID, people are more likely to visit their local or personal general practitioner for the mild covid symptoms. if the symtoms develop into long covid and need to be hospitalized, then thats 
% - more data on how the initial stages of covid infections infect the patients can be studied
% - more research can be conducted to devise treatment plans proper care on others. 
% - REACH A COMMON GOAL
% - }
% In this paper, we describe how traditional data-sharing approaches relying upon conventional privacy-enhancing technologies are limited by various regulations governing medical use and data sharing.

Data sharing can enhance the process of devising treatment plans and diagnosing patients. Furthermore, collecting data from multiple facilities allows scientific breakthroughs to reach another level beyond those obtained from one research. This study evaluates the classification and prediction performance of our multi-site split learning algorithm whilst varying the number of clients and the data ratio. 

The results of this study illustrate that medical data can be shared amongst other institutions whilst preserving privacy. From testing with COVID-19 chest CT scans, MURA X-ray images, and cholesterol levels, a correlation between the number of end-systems and the split ratio is formed. As mentioned in the results section, it can be seen that when the data ratio varies from being equally distributed, to having a data imbalance, to extreme data imbalance, the accuracy levels gradually increase. In our experimental setup, when we say that the data is equally shared amongst the participating hospitals, it is analogous to small hospitals with small amounts of data training one neural network. Notice the accuracy levels in Table~\ref{tab:coco}, \ref{tab:muraacc}, and \ref{tab:chol} is at its peak when the split learning ratio is divided such that one hospital holds a large portion of the data and the other hospitals equally share the remaining data. In other words, in the case where there are 3 end-systems, the best accuracy is seen when the data is divided in a 8:1:1 ratio for all seven of the body portions. The same analysis can be made with the cases of 4 end-systems and 5 end-systems; the highest accuracy level is achieved with a split ratio of 7:1:1:1 and 6:1:1:1:1, respectively. 

The reasoning behind this trend becomes more comprehensible when a simple yet logical comparison to a realistic setting is made. When several small private clinics with a lower number of patient records work only with each other to train a deep neural network, it will obviously perform less. However, when an extremely large-scale hospital, such as The Johns Hopkins Hospital or Massachusetts General Hospital joins that group of small clinics in split learning, then more accurate classifications or predictions will be made. 
Since small local clinics don't accumulate as much data as extensive teaching hospitals, a performance difference is bound to occur. Therefore, having an equal split of data whilst one hospital holds on to the majority of the data, gives the most accurate classification for medical image data. The same pattern is seen across all the datasets tested in this paper. 

Theoretically, when the datasets are equally shared among the participants, the problem of internal covariate shift occurs. In other words, since the batch sizes are the same, during training the solution may reach local optimum rather than the global optimum. However, when a large dataset joins the other small groups (i.e., 7:1:1:1), then the distribution of inputs flowing throughout the network shifts to be around the same mean and standard deviation. Thereby, giving a higher performance rate. The results of our study are corroborated both logically and theoretically. 

The experiments conducted throughout this study allows us to give empirical verification on the optimal number of end-systems and split ratio. It is finalized that for optimal performance for split learning is to have 5 end-systems with the split ratio being 6:1:1:1:1. To achieve the most efficient classification or prediction performance under split learning, research should be conducted with 5 hospitals where one major hospital with the majority of the data joins four other small clinics that have equal amounts of data among themselves. This conclusion is made based on the consideration of both the split ratios and end-systems. This is a guideline to make harmonized and concerted efforts for those in the medical environment wishing to create more meticulous healthcare developments. 

This optimal configuration does not give the highest classification accuracy or lowest loss. The arrangement that gives the best performance is when there are 3 end-systems with a split ratio of 8:1:1. However, we select our proposed composition to be the optimal choice because the difference in accuracy levels is only 0.9\%. By sacrificing a drop in performance of less than 1\%, more hospitals are encouraged to engage in split learning and share their findings to create substantial medical findings. Thereby, forgoing this infinitesimal decline in performance gives smaller hospitals the opportunity to participate in training a common deep neural network. 

\typeout{Our methods allow a transparent use of data, especially in current medical trends where collecting of medical data is on the rise and possessing big data creates paths for scientific advancement. The benefit of split learning in regards to privacy and overcoming data-imbalance can gain consumer trust and willingness to share more information to create meaningful medical deductions. Creating a larger aggregation of datasets will be able to provide a holistic view of emerging health problems. We hope that more partnerships between small local clinics and major teaching hospitals can be formed for fundamental improvements in the medical field to occur. }

\section*{Conclusion}
% I recommend a separate Conclusion section where the researchers 
% -wrap up and confirm the outcomes of the study
% -provide their significance and implications. 
% -The limitations of the research should be stated
% -suggestions for future research direction on the topic. 
% -Concrete recommendations 
 To provide a solution to the practice of maintaining the confidentiality of patient information, we introduced multi-site split learning. Through cross-disciplinary  experiments, we provided the most optimal data ratio and number of end-systems the users may take to achieve ideal performance. This algorithm allows various hospitals to share their data with other institutions without fear of exposing sensitive patient records. 
Our methods allow a transparent use of data, especially in current medical trends where collecting of medical data is on the rise and possessing big data creates paths for scientific advancement. The benefit of split learning in regards to privacy and overcoming data-imbalance can gain consumer trust and willingness to share more information to create meaningful medical deductions. It is recommended when using our multi-site split learning algorithm, the users go through an image pre-processing step to ensure the patient image data are equal in size. The next step in this research would be to create a larger aggregation of datasets as this will provide a holistic view of emerging health problems. Future work will address more advanced ways to encrypt the original patient data and to include another step that will guarantee security even among the exposed feature maps. We hope that more partnerships between small local clinics and major teaching hospitals can be formed for fundamental improvements in the medical field to occur.

\section*{Data availability}
The COVID-19 and MURA data generated or analyzed during this study are included in this published article. 
The cholesterol data is a privately obtained dataset from SNUH is made available from the corresponding author upon reasonable request. Our source codes for the current study are available at https://github.com/annabana17/NatureSplitLearning.

\bibliography{sample}

% \noindent LaTeX formats citations and references automatically using the bibliography records in your .bib file, which you can edit via the project menu. Use the cite command for an inline citation, e.g.  \cite{Hao:gidmaps:2014}.

% For data citations of datasets uploaded to e.g. \emph{figshare}, please use the \verb|howpublished| option in the bib entry to specify the platform and the link, as in the \verb|Hao:gidmaps:2014| example in the sample bibliography file.

\section*{Acknowledgements}

This study was approved by the Institutional Review Board of Seoul National University Hospital (No. C-1712-009-903) with a waiver of informed consent. No personally identifiable data was included in the dataset. Data used in this study were retrieved from Seoul National University Hospital's Common Data Model (CDM) database. 

This research is funded by the National Research Foundation of Korea (2019R1A2C4070663).

\section*{Author contributions statement}
% Conceptualization:
% Methodology:
% Sofware development:
% Writing:
% Supervision:
% Project administration:
% Funding acquisition: 
Y.J.H  designed and conducted the experiments, analyzed the data, prepared figures, and wrote the manuscript; G.L. and M.J. aided in the experiments and plotted the graphs; J.K. supervised the research; S.J., S.Y, J.K. and Y.J.H. reviewed and approved the final version of the manuscript. 

% Must include all authors, identified by initials, for example:
% A.A. conceived the experiment(s),  A.A. and B.A. conducted the experiment(s), C.A. and D.A. analysed the results.  All authors reviewed the manuscript. 

\section*{Competing interests}
The authors declare no competing interests.

\section*{Additional information}

To include, in this order: \textbf{Accession codes} (where applicable); \textbf{Competing interests} (mandatory statement). 

The corresponding author is responsible for submitting a \href{http://www.nature.com/srep/policies/index.html#competing}{competing interests statement} on behalf of all authors of the paper. This statement must be included in the submitted article file.

\end{document}